# *TrackGPT* – A generative pre-trained transformer for cross-domain entity trajectory forecasting

**Nicholas Stroh**
Illumina Consulting Group
`nick.stroh@{icgsolutions.com, gmail.com}`

*Abstract* – The forecasting of entity trajectories at future points in time is a critical capability gap in applications across both Commercial and Defense sectors. Transformers, and specifically Generative Pre-trained Transformer (GPT) networks have recently revolutionized several fields of Artificial Intelligence, most notably Natural Language Processing (NLP) with the advent of Large Language Models (LLM) like OpenAI's ChatGPT. In this research paper, we introduce TrackGPT, a GPT-based model for entity trajectory forecasting that has shown utility across both maritime and air domains, and we expect to perform well in others. TrackGPT stands as a pioneering GPT model capable of producing accurate predictions across diverse entity time series datasets, demonstrating proficiency in generating both long-term forecasts with sustained accuracy and short-term forecasts with high precision. We present benchmarks against state-of-the-art deep learning techniques, showing that TrackGPT's forecasting capability excels in terms of accuracy, reliability, and modularity. Importantly, TrackGPT achieves these results while remaining domain-agnostic and requiring minimal data features (only location and time) compared to models achieving similar performance. In conclusion, our findings underscore the immense potential of applying GPT architectures to the task of entity trajectory forecasting, exemplified by the innovative TrackGPT model.

## 1 Introduction

The task of predicting the future paths or movements of individual entities (vessels, vehicles, aircraft, individuals) is of critical importance to several industries, including defense (maritime domain awareness, air domain awareness, activity-based intelligence), finance (commodities trading), transportation (public transportation planning), and more. Accurate and reliable forecasting of entity trajectories has far-reaching applications, including improved safety, efficiency, and strategic decision-making processes [17].

In recent years, the application of deep learning techniques to trajectory forecasting has shown remarkable promise. Models like Recurrent Neural Networks (RNNs), Long Short-Term Memory networks (LSTMs), and Convolutional Neural Networks (CNNs) have demonstrated the capability to capture temporal dependencies and spatial information in trajectory data. However, the state-of-the-art in this field remains an evolving domain, with several challenges, including the failure of models to generalize across domains (e.g., maritime vs air), inaccuracy of long-term forecasts [7], and dependence on unreliable and/or unavailable features in the data.

Transformer networks [1] have become synonymous with state-of-the-art natural language understanding and generation. Their inception marks a significant milestone in the history of AI and NLP. The versatility of these networks is showcased through their applications in various domains; from chatbots and virtual assistants to content generation, translation, summarization, and even medical diagnosis [18].

In the realm of Natural Language Processing (NLP), transformer-based language models have revolutionized the way machines understand and generate human language. The traditional transformer model consists of two main components: an encoder and a decoder. However, OpenAI's GPT models have deviated from this norm by adopting a decoder-only architecture. This means that the input data is fed directly into the decoder without being transformed into a higher, more abstract representation by an encoder.

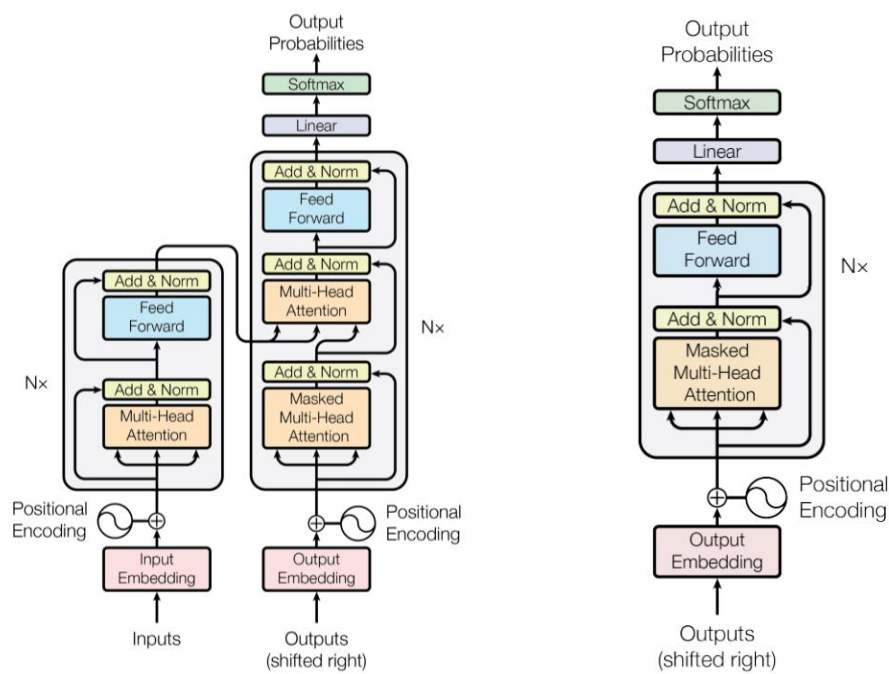

**Fig. 1.** Transformer architectures **Left:** Encoder/Decoder transformer architecture [1] **Right:** Decoder-only GPT architecture.

The decoder in a GPT model uses a specific type of attention mechanism known as masked self-attention. In a traditional transformer, the attention mechanism allows the model to focus on all parts of the input when generating each part of the output. However, in a decoder-only transformer like GPT, the attention mechanism is "masked" to prevent it from looking at future parts of the input when generating the output. This is necessary because GPT models are trained to predict the next token in a sequence, so they should not have access to future tokens during the training process.

The GPT decoder-only architecture simplifies the model and makes it more efficient for certain tasks, like language modeling. By removing the encoder, GPT models can process input data more directly and generate output more quickly. This architecture also allows GPT models to be trained on a large amount of unlabeled data, which is a significant advantage in the field of NLP where most large training datasets are unlabeled.

In this paper, we introduce "TrackGPT," a GPT model designed for the task of cross-domain entity trajectory forecasting rather than NLP. Building upon the success of the GPT architecture in various natural language processing and computer vision tasks, TrackGPT leverages the power of self-attention mechanisms to capture dependencies in geospatial data. In contrast with other forecast models which require a multitude of features from datasets (speed over ground (SOG), course over ground (COG), wind speed and direction, etc.), TrackGPT was designed to require the minimum of data features for trajectory forecasting, only position and time. This flexibility positions TrackGPT as a portable model architecture which can generalize widely without code modification. Despite these minimal data requirements, we will show TrackGPT outperforming models in standardized benchmarks which were purpose-built for their datasets, with more complex architectures and with access to additional features of the dataset.

In the following sections, we provide a detailed description of the TrackGPT model, the experimental setup, and results. We conclude with a discussion of the broader implications of our work and directions for future research in the field of trajectory forecasting.

## 2   Problem Statement and Related Work

The problem TrackGPT aims to solve is to use historical track data consisting only of latitude, longitude, and timestamps to forecast the position of entities at specified future points in time. We denote an observation consisting of latitude and longitude at time $t$ by $x_t$. TrackGPT interpolates raw historical observation data $x_{0:T} \triangleq \{x_{t0}, x_{t1}, \ldots, x_{tT}\}$ into a set of observation data at fixed time intervals $dt$, $\{x_{t0}, x_{t0+dt}, x_{t0+2dt}, \ldots x_{t0+ndt}\}$ where $t_T - dt < ndt <= t_T$. The value of $dt$ is set as the maximum value between the minimum $dt$ that allows tracks to fit in a single block ($dt_{mc}$) and the maximum value of dt that will allow all track geohashes to be neighbors (or the same) of previous values ($dt_{an}$): $dt = \max(dt_{an}, dt_{mc})$. Given the interpolated historical data, we aim to forecast future positions at each future $dt$ increment until a specified future time L, $x_{T+1:T+L}$, so we estimate the following conditional probability distribution:

$$p(x_{tT+1}:x_{tT+L} \mid x_{0:T})$$

Historically, time-series forecasting has relied on human domain knowledge to define parametric models such as exponential smoothing, autoregressive, or structural time series models [21]. The advent of deep learning models has greatly reduced, and often eliminated the requirement for domain knowledge, allowing models "to learn chronological dynamics purely from data" [21]. Until recently, the state-of-the-art in deep learning application to entity trajectory forecasting has been Recurrent Neural Networks (RNNs), specifically Long Short-Term Memory (LSTM) networks [13][14][15][16]. "The four interacting layers of a repeating module in an LSTM enables it to connect the long-term dependencies to present predicting task. Applying sliding windows in LSTM maintains the continuity and avoids compromising the dynamic dependencies of adjacent states in the long-term sequences, which helps to improve accuracy of trajectory prediction" [15]. Many of these forecasting models employed sequence-to-sequence (Seq2Seq) architectures, where input time series data was encoded to context vectors, which are then decoded into forecasts. Non-transformer Seq2Seq models such as Gated Recurrent Unit (GRU) networks have been successfully applied to vessel forecasting [22]. Since the release of OpenAI's GPT-3 in 2020, researchers have begun successfully applying Transformer architectures [1] to entity trajectory forecasting. Applications have been made in forecasting short-term pedestrian trajectories [10][11], as well as long-term vessel trajectories [7]. We are not aware of other Transformer architectures that have been applied across entity domains before TrackGPT. As of 2023, GPT models have been successfully applied to time series forecasting [12], though we are not aware of other GPT-based models used in entity trajectory forecasting.

## 3   Proposed Approach

In this section we detail the unique components used in our approach to forecasting future entity positions by processing their prior and current positions. We introduce a novel, single-dimensional representation of entity position data, and describe the functions of TrackGPT's Geospatial Encoder, Geospatial Decoder, and Forecast Regulator.

*3.1 Input Representation*

In the realm of GPT models, the selection of appropriate data representation methods, specifically token embeddings, is crucial for achieving high levels of performance and accuracy. Token embeddings are the way raw text data is transformed into a format which neural networks can process. The choice and quality of these embeddings directly influences the model's ability to understand and contextualize the data [19]. The choice of embeddings also impacts the model's efficiency and scale, as it determines the dimensionality and complexity of the data being processed, which dictates network parameter count. The selection of token embeddings is a critical step that determines the overall efficacy, applicability, and performance of GPT models.

When applied to NLP tasks, it would be computationally infeasible (as of the writing of this paper) to have a token for each unique word, and computationally inefficient (in all but trivial models) to have a token for each letter, so token representations are typically combinations of several characters as a middle-ground. This tokenization choice gives modern GPT implementations vocabulary sizes (the total number of distinct tokens) of roughly 100,000 or less.  In the field of entity trajectory forecasting, our input space is all positions on the globe at arbitrary precision levels.  An example of a prior technique for encoding position data is to choose a fixed resolution for latitude and longitude, e.g. .01 degree [7].  These fixed resolutions are often combined with other data features such as course and speed, resulting in complex multidimensional data representations. "The most common way to represent an AIS message is using a 4-D real-valued vector (two dimensions for the position and [the] other two for the velocity, e.g. [lat,lon,SOG,COG]$^T$" [24]. We propose a more straightforward and flexible approach in our Geospatial Encoder module: representing the trajectory data simply as an ordered sequence of geohashes.

Geohashing offers a convenient method of representing locations with arbitrary degrees of precision, using a hierarchical, grid-based approach. The geohash system divides the Earth's surface into a grid of 32 rectangles, and each subsequent character in a geohash represents a further subdivision of one of these rectangles into 32 parts, allowing for progressively finer spatial resolution.

In our initial implementation of TrackGPT, we used 16-bit unsigned integers as our input data type, giving us a maximum vocabulary size of 65,536. This modest vocabulary size was ideal for our available testing hardware[1]. Each geohash character requires 5 bits, allowing us to fully represent 3 geohash characters and have an extra bit left over. We chose to use this bit as an east/west flag within the geohash box, doubling the precision of the input (and output) representation. For the remainer of this paper, we'll refer to this subdivision of a geohash as ".5", so "3.5" character geohash resolution means half of a 3-character geohash.

In the field of entity trajectory forecasting, 3.5-character resolution would be insufficient for most problem sets, yielding a 78km precision level. To solve this problem, our Geospatial Encoder leverages the hierarchical nature of geohashes and stores a fixed prefix for each geohash, representing the geospatial extent of the area being modeled. TrackGPT may also shift datasets to neighboring geohashes on input if it results in a larger fixed geohash prefix and therefore increased forecast resolution. Any shift performed on input is automatically stored and reversed on the forecast output. This approach allowed us to process 5-character data (2.4km precision) in our long-term vessel forecasting benchmark (Section 4.1), and 8-character data (0.019km precision) in our short-term aircraft forecasting benchmark (Section 4.2).

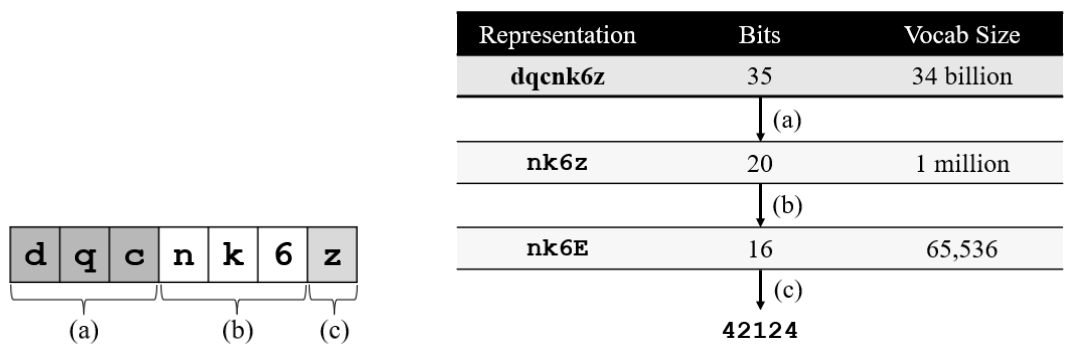

**Fig. 2.** TrackGPT input encoding **Left:** Geohash character representation. (a) Prefix stored in metadata. (b) Encoded full precision. (c) Encoded as a 1-bit flag. **Right:** Compressing a 35-bit geohash into a 16-bit vocabulary. (a) Store "dqc" as a static prefix for the model. (b) Convert the last character to "E" or "W" flag (c) map to int16 vocabulary.

*3.2 Input Sampling*

The interpolation and sampling interval of input data is of crucial importance to training a successful model. Entity track data at scale is rarely sampled at uniform intervals in the temporal or geospatial domain, and blackout periods are common in otherwise uniform tracks [20]. Since the input to TrackGPT is purely vectors of geospatial positions, uniform sampling allows the model to implicitly learn other features of the track (e.g., course, speed) and project the track forward to specific future time intervals. The sampling interval of the interpolated tracks should be high enough such that as many of the representative geohashes in successive points are neighbors as possible, allowing the model to quickly learn the geospatial relationship between neighboring tokens. Lower sampling intervals where tracks jumped several geohashes at a time yielded poor model performance.

---
[1] 12-core cloud VM, single NVIDIA A100 80GB GPU

To provide the network with tracks suited to GPT training, the Geospatial Encoder module grooms raw track data in the following ways:

- Splits tracks with blackout periods over a configurable amount of time
- Removes tracks under a configurable duration
- Interpolates tracks
- Re-samples tracks at a configurable fixed interval

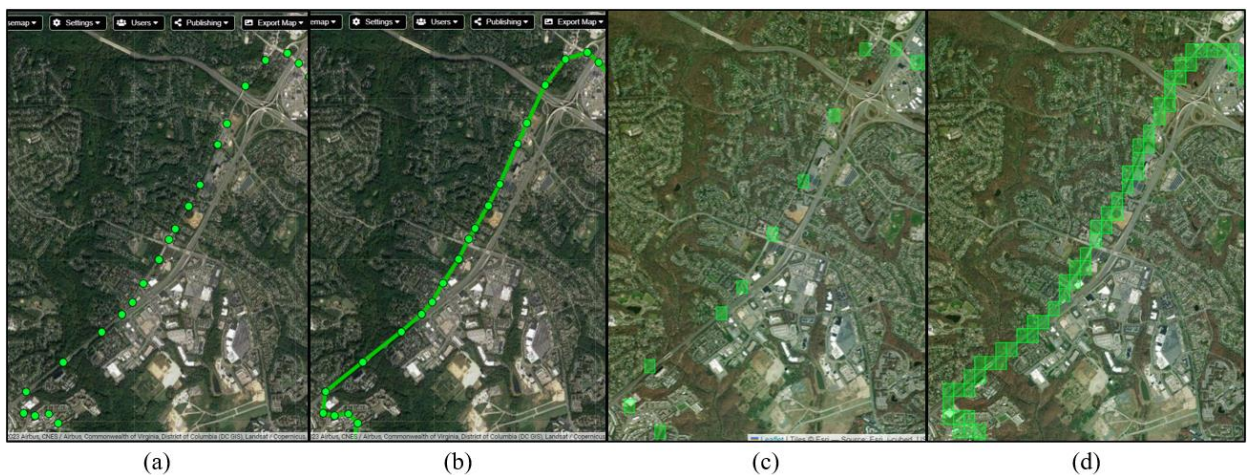

(a)      (b)      (c)      (d)

**Fig. 3.** TrackGPT interpolation and sampling of raw track data: **(a)** Raw track data input, **(b)** Interpolated track, **(c)** Track sampled at low frequency leads to geohash "jumps", not ideal for network training, **(d)** Track sampled at higher frequency leads to contiguous geohash tracks.

*3.3 Forecast Regulator*

The TrackGPT architecture includes a Forecast Regulator module designed to enhance and ensemble forecast predictions as well as automatically eliminate hallucinations.

"Artificial hallucination refers to the phenomenon of a machine, such as a chatbot, generating seemingly realistic sensory experiences that do not correspond to any real-world input. This can include visual, auditory, or other types of hallucinations. Artificial hallucination is not common in chatbots, as they are typically designed to respond based on pre-programmed rules and data sets rather than generating new information. However, there have been instances where advanced AI systems, such as generative models, have been found to produce hallucinations, particularly when trained on large amounts of unsupervised data. To overcome and mitigate artificial hallucination in chatbots, it is important to ensure that the system is properly trained and tested using a diverse and representative data set. Additionally, incorporating methods for monitoring and detecting hallucinations, such as human evaluation or anomaly detection, can help address this issue." [2]

To eliminate hallucinations, the Forecast Regulator examines each forecast output by the TrackGPT decoder. Since the Geospatial Encoder ensures tracks contain few geohash gaps, the Forecast Regulator can simply check that successive predictions are no more than N hops apart, where N is an integer hyperparameter set to be appropriate for the dataset. In our test datasets we found N=3 for vessels and N=10 for aircraft yielded the best results. We attribute this difference both to the speed difference between entity types and the different levels of precision used in the tests (a faster moving entity over smaller geohash bounds will produce geohash gaps more frequently compared to a slower moving entity with larger geohash bounds). If gaps of more than N are found, forecast output is truncated at that point.

TrackGPT leverages batch predictions to perform multiple simultaneous forecasts per input track. The optimal model temperature[2] for forecast variation and low hallucinations rate was determined to be 0.92. The Forecast Regulator can then ensemble N forecasts and determine mean routes, consensus destinations, and waypoints to include in final output. "The ability of ensemble systems to improve deterministic-style forecasts and to predict forecast skill has been convincingly established. Statistically significant spread-error correlations suggest that ensemble variance and related measures of ensemble spread are skillful indicators of the accuracy of the ensemble mean forecast" [3]. Forecast Regulator functionality is tunable to the problem set via hyperparameters.

*3.4 Architecture*

The core of TrackGPT is implemented using Andrej Karpathy's NanoGPT library [4], which is used to instantiate a 792 million parameter GPT model. NanoGPT's data loader is designed for NLP tasks, where random blocks of text and punctuation tokens are sampled during training. To make the data loader suitable to track input, we've modified the sampling logic to extract blocks only from within a single track at a time (starting from a randomized offset), and not "wrap" sampling to subsequent tracks. In this way, we ensure coherent input vectors.

---

[2] Temperature in generative models is a hyperparameter that affects the randomness of the model output. Temperatures closer to 0 make the model more deterministic, while higher temperatures of 1 and above produce less predictable output.

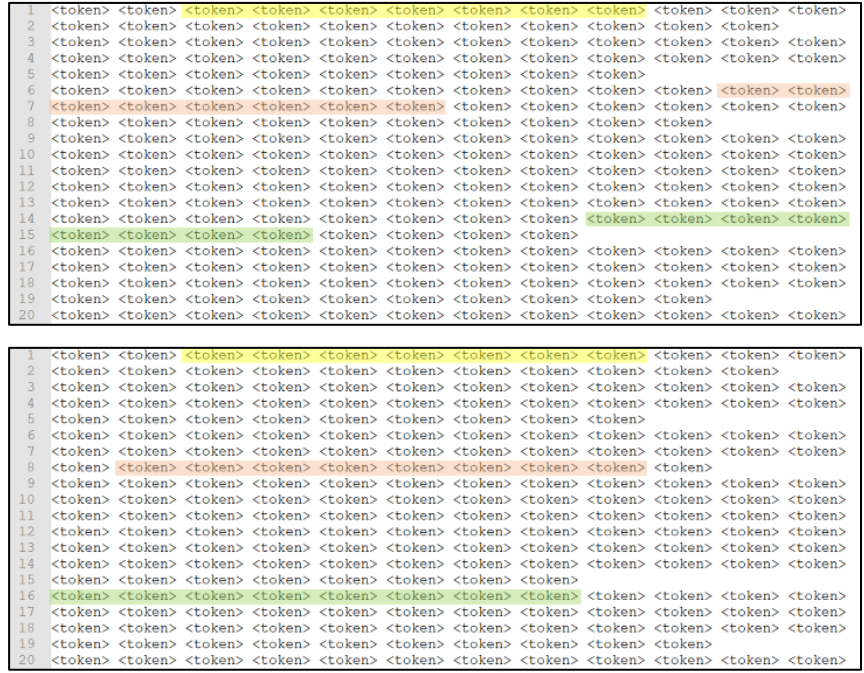

**Fig. 4.** Difference in data loading methodologies between NanoGPT and TrackGPT. **Top:** NanoGPT data loader selects random blocks of text data, line breaks are treated as tokens **Bottom:** TrackGPT loads blocks from random offsets within contiguous tracks which are newline-delimited, ensuring the network isn't trained on fragments from multiple tracks.

Raw track data is fed into the Geospatial Encoder, which performs the functions outlined in Sections 3.1 and 3.2 to prepare the data for the GPT network. The output of the GPT network goes to a Geospatial Decoder component which decodes the token output into geohash tracks. At this stage, geohashes are re-assembled with any stored fixed prefix and geohash shifting used to fit the data into a 16-bit vocabulary. The Forecast Regulator processes output from the Geospatial Decoder and performs the functions outlined in Section 3.3 before producing the final forecasts.

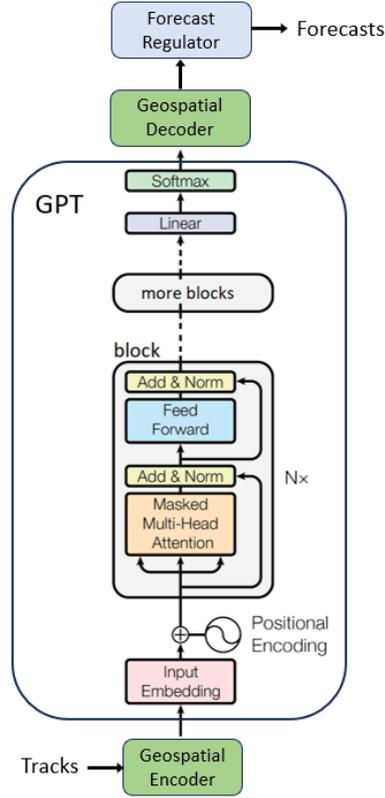

**Fig. 5.** TrackGPT architecture diagram.

## 4   Results

In this section we evaluate TrackGPT's forecast performance by benchmarking it against two public datasets with published results: Danish Maritime Authority (DMA) Automatic identification system (AIS) data [5] and TrajAir Automatic Dependent Surveillance-Broadcast (ADS-B) data [6]. We chose these datasets to test TrackGPT's applicability in both long- and short-term forecasting, at different levels of precision, and to both the maritime and air domains. In both tests, TrackGPT used only location and time features in the data, where other benchmarked models used additional features, were often custom-built networks for the dataset, and could not be applied to other domains [8].

*4.1 Long-term Vessel Forecasting*

TrackGPT was benchmarked against public AIS data provided by the DMA [5] following the methodology outlined in *TrAISformer* [7]. The dataset consists of AIS records within the DMA Area of Interest (AOI) collected between January 01, 2019 and March 31, 2019. The dataset contains over 700 million AIS records, and was split into training (January 01, 2019 to March 20, 2019) and test (March 21, 2019 to March 31, 2019) sets.

Methodology [7]:
- Removal of stationary tracks (all positions resolve to the same geohash)

- Split tracks with an interval between AIS records of greater than 1 hour
- Removal of tracks with duration under 4 hours
- Interpolate and re-sample tracks to 10-minute intervals
- Split long duration tracks so the maximum duration for any track is 20 hours

The raw AIS records were converted into 5-character geohash sequences. Data was preprocessed as outlined above and used for model pre-training using our modified data loader. The testing dataset consisted of all non-stationary 20-hour track segments between March 21, 2019 and March 31, 2019. The goal of the benchmark was to forecast tracks up to 20-hours in duration [7], so our data loader was configured to load 20-hour blocks during training. For each track in the testing set, the first 5 hours of each track was used as a prompt to the model, and the model output predictions at 10-minute intervals for 15 hours:

$$p(x_{tT+1}:x_{tT+90} \mid x_{tT-30}:x_{tT})$$

Following the benchmark methodology [7], best-of-N criterion was used to score predictions, where N = 16. Prediction error was measured at 1-hour intervals using the geodesic distance between predicted and actual track locations. Where actual locations fell within the predicted geohash bounding box, error was recorded at 0, where it was outside of the bounds error was recorded as distance to the closest corner of the predicted bounding box. Given our available hardware and the size of the benchmark AOI, the data was run with 5-character geohash precision, resulting in 1.3NM resolution in the benchmark AOI, and forecasts were output at 4.5 character resolution.

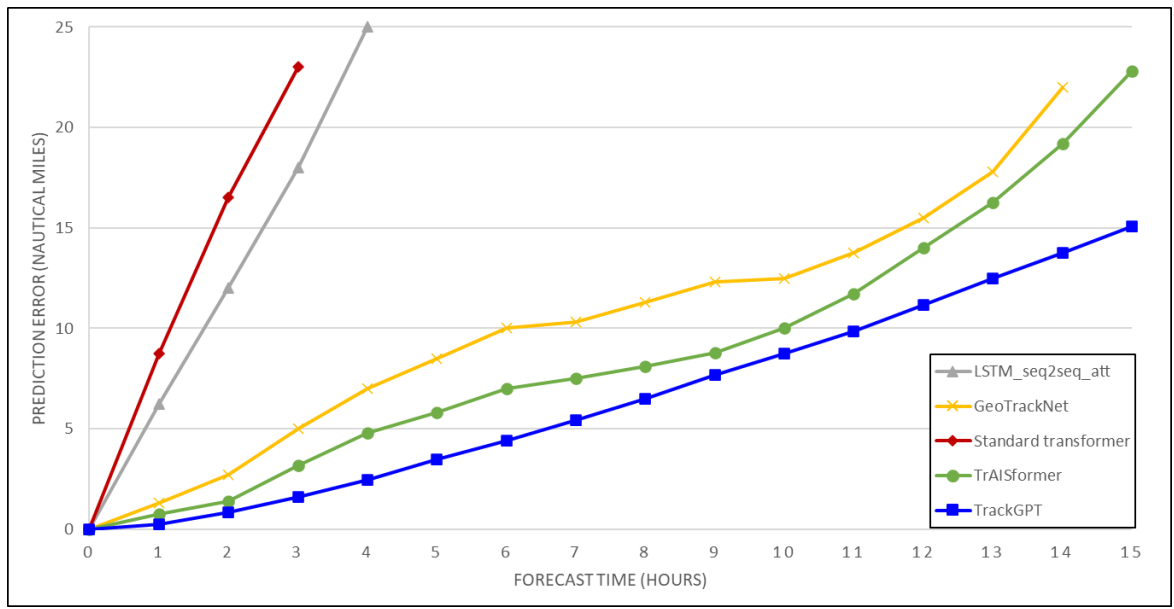

**Fig. 6.** DMA AIS benchmark results for TrackGPT vs other models [7] showing forecast error (NM) vs forecast time (h) for each model.

**Table 1.** TrackGPT DMA AIS benchmark results at each 1-hour forecast interval.

| Forecast Time (h) | 1 | 2 | 3 | 4 | 5 | 6 | 7 | 8 | 9 | 10 | 11 | 12 | 13 | 14 | 15 |
|---|---|---|---|---|---|---|---|---|---|---|---|---|---|---|---|
| Forecast Error (NM) | 0.25 | 0.86 | 1.6 | 2.47 | 3.46 | 4.42 | 5.43 | 6.5 | 7.67 | 8.74 | 9.83 | 11.15 | 12.48 | 13.74 | 15.07 |

Despite only using time and location features, TrackGPT produced the most accurate forecasts at every time interval compared with all other models' published benchmarks [7]. 15-hour forecasts were on average 7.5NM more accurate than the next-best model. It should be noted that other models had better resolution in their forecasts [7]; *TrAISformer* performed within a 1.3NM margin of error at 1-hour and 2-hour evaluation periods. Using a 6-character geohash would have been ideal for this benchmark, providing 0.33NM resolution, however given the size of the benchmark AOI this results in a network too large to train on our available single-GPU testing hardware.

TrackGPT performed exceptionally well compared to other models at longer term forecasts of 10h-15h, likely due to the model being trained with a 20-hour context window and encoding long-term associations between geohashes. We especially note the performance improvements of TrackGPT versus the two other transformer-based models that have been benchmarked, the Standard Transformer [1] and TrAISformer [7]. One possible explanation for this delta is that TrackGPT doesn't rely on SOG or COG features of the dataset, which are notoriously unreliable in AIS (and similar) datasets, and instead lets the model implicitly determine these features solely on the series of positions. "Static and dynamic AIS data, received and recorded on a data carrier, are occasionally false or incomplete, as they contain empty fields … reliability limitations may result from the incorrectly estimated course over ground (COG) or speed over ground (SOG), falsely describing the dynamics of the vessel" [23]. Another distinguishing feature of TrackGPT is the fully autoregressive nature of GPT architectures, compared to encoder/decoder architectures in which the encoder step is not autoregressive. Autoregressive models have a long history of success in time series forecasting [25], so this may be an advantage over traditional transformer architectures, where the encoded input context is static.

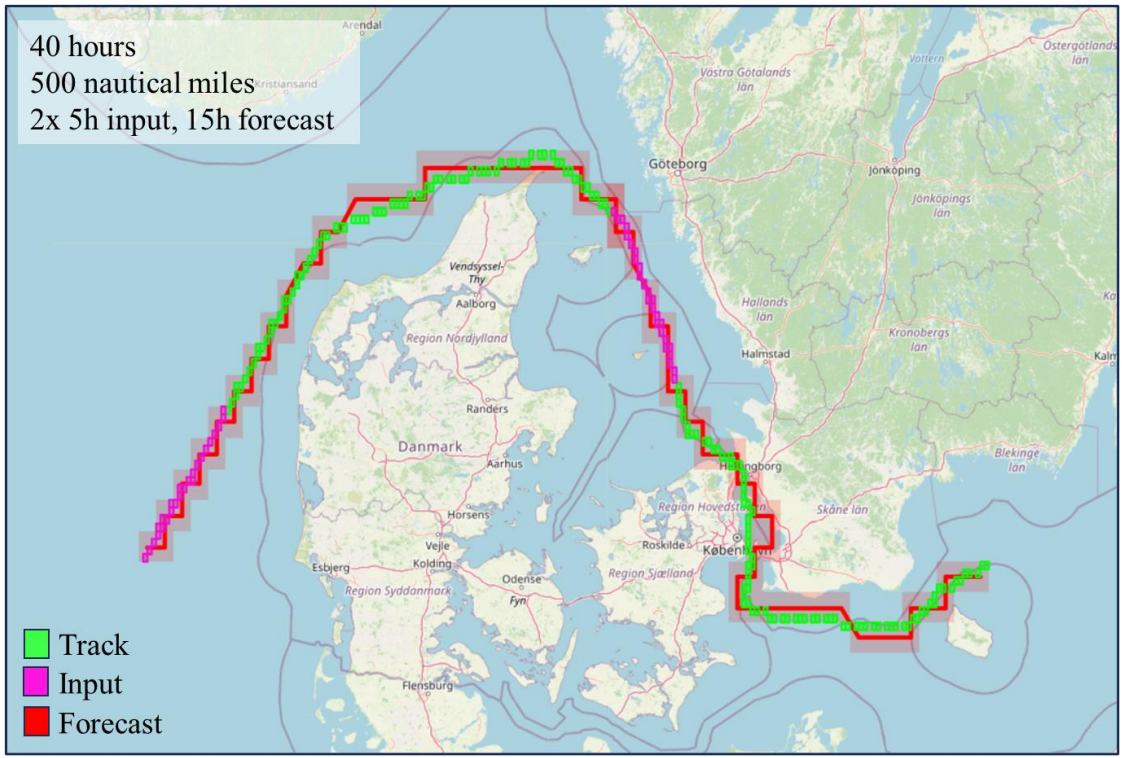

**Fig. 7.** Back-to-back TrackGPT forecasts of a vessel over a 40-hour, 500 nautical mile voyage from the DMA AIS benchmark dataset. Each 15-hour forecast was made using 5-hours of input data.

*4.2 Short-term Aircraft Forecasting*

TrackGPT was benchmarked against the public TrajAir ADS-B dataset [6] following the methodology outlined in the dataset's associated research paper [8].

"The TrajAir dataset is collected at the Pittsburgh-Butler Regional Airport (ICAO:KBTP), a single runway GA [General Aviation] airport, 10 miles North of the city of Pittsburgh, Pennsylvania … The trajectory data provided spans days from 18 Sept 2020 till 23 Apr 2021 and includes a total of 111 days of data discounting downtime, repairs, and bad weather days with no traffic … The dataset uses an Automatic Dependent Surveillance-Broadcast (ADS-B) receiver placed within the airport premises to capture the trajectory data." [6]

The TrajAir dataset consists of both raw and pre-processed versions of the ADS-B data [6]. Since TrackGPT was designed as a cross-domain model applicable to any geospatial dataset and sampling and interpolation are part of the architecture, the unprocessed raw dataset was used in our tests. While other models benchmarked had large architectural components dedicated to processing weather data and other features specific to this dataset [8], TrackGPT used only time and location features. No architectural changes were required for TrackGPT to process ADS-B data versus AIS data.

Methodology [8]:

- Removal of duplicate records for the same aircraft identifier and time fields

- Removal of data points where the altitude is greater than 6000 ft
- Removal of data points where the distance from the runway is greater than 5km
- Interpolate and re-sample tracks to 250 millisecond intervals
- Split long duration tracks so the maximum duration for any track is 120 seconds

The raw ADS-B records were converted into 8-character geohash sequences. Data was preprocessed as outlined above and used for model pre-training using our modified data loader. The testing dataset consisted of all 120 second or longer track segments between November 26, 2020 and November 29, 2020. The goal of the benchmark was to forecast tracks up to 120-seconds in duration [8], so our data loader was configured to load 120-second blocks during training. For each track in the testing set, the first 11 seconds of each track was used as a prompt to the model, and the model output predictions at 250 millisecond intervals for 109 seconds:

$$p(x_{tT+1}:x_{tT+436} \mid x_{tT-44}:x_{tT})$$

Following the benchmark methodology [9], best-of-N criterion was used to score predictions, where N = 5. Average Displacement Error (ADE) was calculated by measuring the geodesic distance between each forecast point $x_{tT+1}:x_{tT+436}$ and the true interpolated track location at the corresponding time, and taking an average. Final Displacement Error (FDE) was calculated by measuring the geodesic distance between the final predicted location at $x_{tT+436}$ and the final location of the track at the corresponding time. Where actual locations fell within the predicted geohash bounding box, error was recorded at 0, where it was outside of the bounds error was recorded as distance to the closest corner of the predicted bounding box. The data was run with 8-character geohash precision, resulting in 30-meter resolution in the benchmark AOI, and forecasts were output at 7.5 character resolution.

**Table 2.** TrajAir benchmark ADE and FDE results. [8]

| Algorithm | ADE (km) | FDE (km) |
|---|---|---|
| Const. Vel | 1.86 | 4.21 |
| Nearest Neigh. | 2.76 | 2.49 |
| STG-CNN | 1.26 | 2.50 |
| TransformerTF (lat,lon,alt,time,wind x,wind y) | 1.76 | 4.13 |
| TrajAirNet (lat,lon,alt,time,wind x,wind y) | 0.78 | 1.55 |
| **TrackGPT (lat,lon,time)** | **0.41** | **0.90** |

Prior state-of-the-art performance on this benchmark was achieved by the TrajAirNet model, which was purpose-built for this dataset [8]. The model's architecture "uses Temporal Convolutional Networks (TCNs) to encode the 3-D trajectory. The dynamic weather context (wind vectors) are encoded using Convolutional Neural Networks (CNNs), which are appended to the encoded trajectory. To encode the social context, we use a Graph Attention Network (GAT) that uses attention to combine data from different agents. Finally, we use Conditional Variational Autoencoders (CVAEs) to produce multi-future acceleration commands, which are then used in a forward Verlet integration to produce future aircraft trajectories for all agents." [8].

Compared to other published benchmark results [8], TrackGPT produced by far the most accurate forecasts compared with all other models in terms of both ADE and FDE. Surprisingly, TrackGPT was able to accomplish these results without considering the available weather features in the dataset, which accounted for significant portions of other models in the test [8]. As in our AIS benchmark, we again posit in this benchmark that the autoregressive nature of TrackGPT's architecture versus TransformerTF's encoder/decoder architecture [11] was advantageous. It is also plausible in this case that adding to the input vector features whose effects on forecast can be inferred by a time series of positions alone (wind x, wind y), has the net effect of confusing the attention mechanism and decreasing model efficacy.

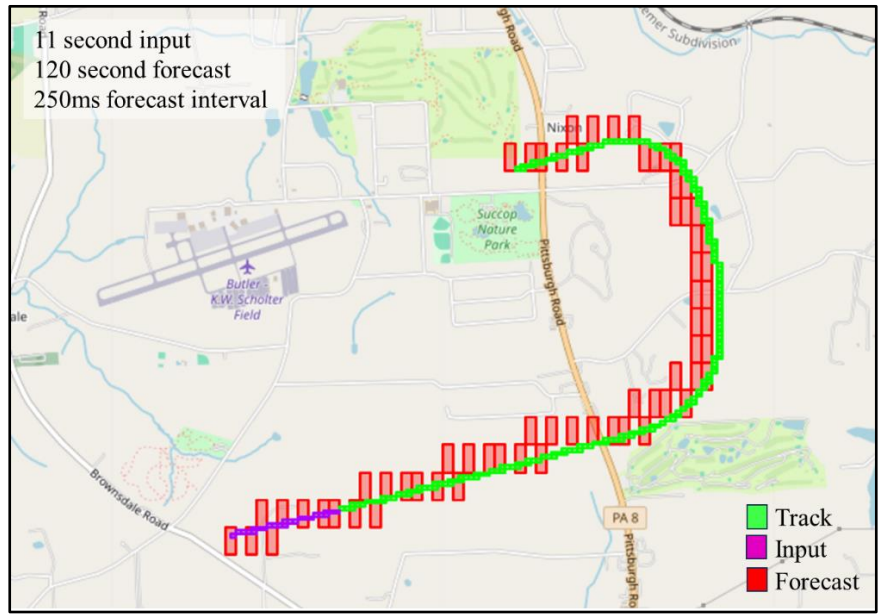

**Fig. 8.** Example TrackGPT forecast result from TrajAir benchmark.

## 5 Conclusions and Future Work

In this work, we present TrackGPT, a generative pre-trained transformer model for cross-domain entity trajectory forecasting. TrackGPT is designed with minimal data feature requirements to ensure applicability across entity domains and datasets. The simple architecture of our solution and no-code portability gives it great potential as a building block for customized domain-specific architectures.

We demonstrated state-of-the-art results on benchmark datasets in two applications: maritime vessel forecasting with AIS and aircraft forecasting with ADS-B. The ability TrackGPT showed in both benchmarks to accurately predict future positions at specific times without considering features such as course or speed, which were used as inputs by competing model, was especially surprising. These attributes seem to have been successfully inferred by the network given inputs of only position and time. One limitation we found in testing was scaling the network sufficiently on our available single-GPU testing hardware; ideally, we could have run the DMA AIS benchmark at 6-character resolution or higher for improved precision.

Given access to a multi-GPU cluster and global dataset, we believe this model would scale well to deliver a global forecast model at precision suitable for many applications in commercial, defense, and financial industries. We posit it may be possible to build this architecture into a single foundational model able to forecast any entity type at a pre-determined level of precision. It is reasonable to assume that given a wide breath of training data across entity types, the model could infer the entity type from an input sequence and forecast accordingly. At the same time we also recognize the opportunity to build customized implementations of this architecture, where dataset-specific layers can be added to leverage additional dataset features, perhaps turning the Forecast Regulator into a plugin architecture for different domains, which should further improve TrackGPT's already-impressive entity trajectory forecast performance.